\documentclass{article}

\usepackage{cite}
\usepackage{amsmath,amssymb,amsfonts}
\usepackage{algorithmic}
\usepackage{graphicx}
\usepackage{textcomp}
\usepackage{xcolor}

\graphicspath{{figures/}}

\def\BibTeX{{\rm B\kern-.05em{\sc i\kern-.025em b}\kern-.08em
    T\kern-.1667em\lower.7ex\hbox{E}\kern-.125emX}}
    
\begin{document}

\title{Investigating the effect of binning on causal discovery\\
{\footnotesize }
\thanks{This work was supported by funding from NCRR 1UL1TR002494-01 to EK.}
}

\author{
Andrew Colt Deckert \\
\textit{Department of Psychology} \\
\textit{University of Minnesota}\\
Minneapolis, USA \\
decke300@umn.edu
\and
Erich Kummerfeld\\
\textit{Institute for Health Informatics} \\
\textit{University of Minnesota}\\
Minneapolis, USA \\
erichk@umn.edu
}

\maketitle

\begin{abstract}
Binning (a.k.a. discretization) of numerically continuous measurements is a wide-spread but controversial practice in data collection, analysis, and presentation. The consequences of binning have been evaluated for many different kinds of data analysis methods, however so far the effect of binning on causal discovery algorithms has not been directly investigated. This paper reports the results of a simulation study that examined the effect of binning on the Greedy Equivalence Search (GES) causal discovery algorithm. Our findings suggest that unbinned continuous data often result in the highest search performance, but some exceptions are identified. We also found that binned data are more sensitive to changes in sample size and tuning parameters, and identified some interactive effects between sample size, binning, and tuning parameter on performance.
\end{abstract}

\newpage
\section{Introduction}
Binning is a pre-processing step widely used in data analysis, where continuous numeric variables are converted into discrete numeric variables. For example, statements such as ``The gap in life expectancy between the richest 1\% and poorest 1\% of individuals was 14.6 years'' makes use of a binned continuous variable, wealth, rather than analyzing it directly as a continuous variable \cite{chetty2016association}. While binning has been extensively discussed in regards to more traditional statistical methods ranging from regression analyses to meta-analysis, the effect of binning on causal discovery algorithms has not been explicitly studied. This paper reviews the common arguments for and against binning, and uses a simulation study to evaluate its effects on a representative causal discovery algorithm. 

The rest of this paper proceeds as follows. Section \ref{background} provides background on the procedure of binning alongside historical and contemporary arguments for and against this procedure, and an overview of the causal discovery algorithm Greedy Equivalence Search (GES). The details of our simulation procedure are described in section \ref{methods}, along with the metrics we used for calculating search performance. In section \ref{results} we report the summarized search performance from those simulations. Lastly, section \ref{discussion} discusses the implications of our findings for optimizing causal discovery search performance, and concludes the paper. 

\section{Background}
\label{background}
\subsection{Greedy Equivalence Search (GES)}
Causal Discovery algorithms seek to learn causal relationships from data. A collection of such relationships is commonly summarized and stored as a graph, and so most causal discovery algorithms are also graph-learning algorithms. Within graphs, variables are stored as nodes or vertices and the causal relationships between them represented by directed edges. Greedy Equivalence Search (GES) is one such causal discovery algorithm \cite{chickering_optimal_2002}. GES searches the space of causal graphs for the one which optimizes a penalized likelihood score, while respecting the existence of equivalence classes of graphs that are statistically indistinguishable.

GES starts with an equivalence class where no dependencies exist between variables and iteratively adds edges, scoring graphs and retaining the better fitting model until no more edges can be added to improve score. The resulting model is then subjected to a similar iterative process of edge deletion retaining the higher scoring model. The original GES algorithm included only forward and backward steps. A third step was proposed where the algorithm executes orientation reversals that result in a higher score \cite{chickering_optimal_2002}. The algorithm for edge reversal was explicated as a step within Greedy Interventional Equivalence Search (GIES). Results show that implementing this turning step not only improves GIES performance but also improves search performance with GES used on observational data as well \cite{hauser2012characterization}. Thus, many newer implementations of GES contain edge reversal.

Implementations of GES, such as that used in the simulations we report here, typically use a decomposable fit score, such as the Bayesian Information Criterion (BIC) score\cite{schwarz197801}, to enhance search and score speed. This allows GES to calculate and compare changes in local scores after a change in edge as opposed to re-scoring the entire graph with each change \cite{chickering_optimal_2002}. Since the BIC score is being used to assess comparative fit between models, rather than absolute fit, the exact formulation is not critical, so long as it is being applied consistently within a single study or algorithm run. Implementations of BIC in some causal discovery packages, such as Tetrad\cite{Tetrad}, also modify the usual definition of BIC by including an additional tuning parameter that can be used to modify the score's degree of preference for simpler models. We make use of such a modification here. So for the purposes of this paper, let the BIC score of model $M$ be:
$$\text{BIC}(M) = -log(L(M)) + \lambda C(M) \log (n) / 2$$
Where $L(M)$ is the likelihood of $M$ given the data, $\lambda$ is a constant real number, $C(M)$ is the complexity of $M$, as determined by the of number of free parameters in $M$, and $n$ is the sample size of the data. $\lambda$ is assigned by the user, and represents the user's degree of preference for simple models with fewer numbers of edges.

GES assumes that cases in the observed data set are i.i.d. and that the underlying probability distribution is faithful to a directed acyclic graph (DAG), a graph with directed edges and no cycles. Algorithm computation time scales with the number of edges in the graph and the maximum number of parents per node.

\subsection{Binning methods}
Binning can come in many forms, some simple and some complex. Binning may be rough by discretizing data into binary categories of ``high'' and ``low''. A common and simple type of discretizing is central tendency splits, such as mean/median splits \cite{MacCallum2002}. However, binning can become increasingly fine-grained by using three or more bins\cite{gelman2009splitting}. Bins may be chosen to have equal width on the continuous measurement scale, or to contain equal numbers of samples, or may be selected by more complicated unsupervised methods such as clustering. Additionally, pre-specified breakpoints may be used based upon theory, convenience or consensus. Supervised binning methods are also available that choose a binning that optimizes predictive information for an outcome of interest. Entropy based binning is one example of a supervised binning algorithm \cite{Fayyad1993}.

\subsection{Arguments for binning}
Binning data was standard practice prior to more advanced computing resources.  Corrections for the inaccuracy of binning were developed for binning with equal width or mean splits \cite{Peters1940}. Today, binning is no longer required as a limitation of computing resources, raising the question of when binning should occur and why? 

Binning today is predominately used for summarizing, grouping, and simplifying data collection, analyses, and visualization. In short, binning is used primarily for parsimony. A common example is dividing Body Mass Index (BMI) into categories of underweight, healthy weight, overweight, and obese. Another commonly seen application of binning in the social sciences is discretization of the extroversion facet of personality into descriptive categories such as introvert and extrovert.

Binning may be used as a form of local smoothing to reduce noise, with the goal of improving signal to noise ratio with an appropriate binning choice. This can preserve the signal but remove small fluctuations in the data, assumed to be noise or outliers. Binning may be used to prepare data for data mining techniques which require discrete values, such as decision trees. Finally, and most importantly for the study presented in this paper, binning may be used to address over-fitting in computer learning algorithms traditionally used on continuous data \cite{MacCallum2002,Friedman1996,Chmielewski2003,Fayyad1993,Frank1999}.

\subsection{Arguments against binning}
The process of binning inevitably results in information loss. The degree of this loss will depend on the number of bins chosen, with dichotomization (bins=2) resulting in the maximal amount of information loss. As a consequence, binning can increase Type I and Type II errors.

Binning is especially likely to lead to Type I errors when the binned variable confounds the relationship between a cause and an outcome. The Type II errors in other paths of the graph leave some of the covariance between the exposure and the outcome unexplained, leading to erroneously inferring that there is an effect from the exposure to the outcome when there is none  \cite{barnwell2015effects}. Type I errors have also been shown to be induced by binning in other ways \cite{MacCallum2002,cohen_cost_1983}.

Dichotomization has been shown to reduce power equivalent to dropping 1/3rd of the sample \cite{MacCallum2002}. The loss of power becomes more prominent the farther the break point is from a central tendency and with the amount of variables which have been binned. Underestimates of effect size are possible \cite{Cohen1983}.

How number of breaks and breakpoints are chosen can also affect results and interpretations. For example, cleverly modifying the break point between ``healthy'' and ``ill'' individuals can lead to an increased life expectancy in both groups.

Binning of conceptually continuous phenomena can occur at any stage of the research process. In cases where binning has occurred at the data collection stage, it is often impossible to recover the lost information.  Researchers have shown that binning affects various statistics such as Pearson’s correlation and both univariate and multivariate analyses \cite{Bollen2006,Cohen1983}.  Other research has shown binning has an influence on  measurement reliability  \cite{Alwin2017,Wainer2006}. Overall, binning can create a misleading and inaccurate picture, but can also create more generalizable and parsimonious results depending on the quality and quantity of the information that is lost (i.e. signal or noise).

\section{Methods}
\label{methods}

\begin{figure}[htbp]
\centerline{\includegraphics[width=.6\linewidth]{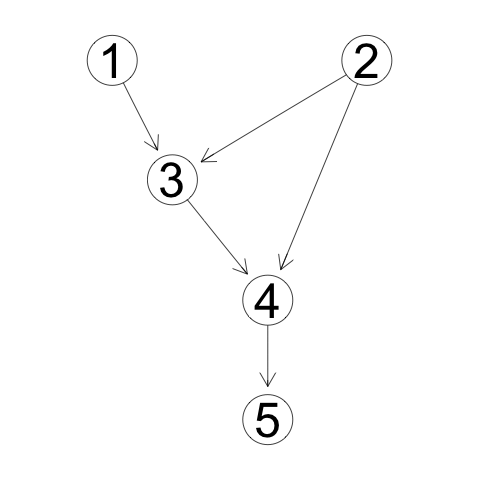}}
\caption{DAG 2}
\label{Dag2}
\end{figure}

\begin{figure}[htbp]
\centerline{\includegraphics[width=.6\linewidth]{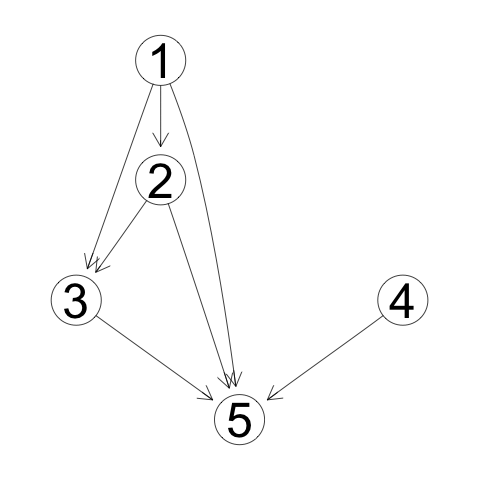}}
\caption{DAG 3}
\label{Dag3}
\end{figure}

The \textit{pcalg} and \textit{graph} packages in R were used for generating the original DAGs as well as following analyses \cite{kalisch_causal_2012,Kalisch2010,Kalisch2012,gentleman2015graph}. Five directed acyclic graphs (DAGs) were generated at random, after fixing their number of nodes and selecting a probability of edge occurrence. Edge weights were then randomly assigned, ranging from 0.1 to 1, and each variable was given an independent noise term sampled from a ~N(0,1) distribution. Data was generated by sampling each variable from a distribution equal to the weighted sum of its parents' values in the model, with weights equal to the corresponding edge weights, and its independent noise term. This process was repeated until the desired number of samples was reached for that data set. Table \ref{gdtab1} contains a summary of metrics on these 5 DAGs, and DAGs 2 and 3 are shown in Figures \ref{Dag2} and \ref{Dag3} respectively. Visualizations of all DAGs can be found in the Online Supplement\footnote{https://github.com/cdecker8/Investigating-the-effect-of-binning-on-causal-discovery-online-supplemental-information}. These DAGs were used to generate the data in our simulations, and served as the ``gold standards'' for evaluating search performance.
\begin{table}[htbp]
\caption{Gold standard DAG metrics}
\begin{center}
\begin{tabular}{|c|c|c|c|c|}
\hline
\textbf{Model}&\textbf{Nodes}&\textbf{Edge Prob.} &\textbf{Edges}  &\textbf{Avg. Degree}\\
\hline
DAG1 & 5 & 0.25 & 3 & 1.20 \\
\hline
DAG2 & 5 & 0.5 & 5 & 2.00  \\
\hline
DAG3 & 5 & 0.75 & 7 & 2.80 \\
\hline
DAG4 & 20 & 0.25 & 51 & 5.10 \\
\hline
DAG5 & 20 & 0.5 & 99 & 9.90 \\
\hline
\end{tabular}
\label{gdtab1}
\end{center}
\end{table}

We generated data sets from each DAG with sample sizes 100, 500, and 1000. 200 data sets were generated from each model at each sample size. Each of these data sets was used to create four additional binned data sets, with 2, 5, 10, and 15 bins of equal interval width, by replacing the continuous data values with corresponding bin values. We also tested the effects of varying the $\lambda$ tuning parameter in our chosen implementation of the BIC score assuming a linear gaussian model. $\lambda$ values tested were 1, 2, and 4. $\lambda=1$ is the standard BIC score, while 2 and 4 increase the score's preference for simpler models with fewer numbers of free parameters, i.e. fewer numbers of edges. This resulted in 15 conditions, each with 200 data sets, for each gold standard DAG.

GES was run on each data set using forward, backward and turning procedures. All data sets were treated as continuous during the scoring process, regardless of binning procedure (or lack thereof), as such discrete BIC, which does not assume a linear Gaussian model was not utilized, only continuous BIC. Rationale for treating all data as continuous include a ``rule of thumb'' that likert scales, frequently used in medicine and related fields, may be treated as continuous at the 5 or 7 response options \cite{sullivan2013analyzing}.

For each data set, the Structural Hamming Distance (SHD), F1 Score (F1), True Discovery Rate (TDR), True Positive Rate (TPR), and False Positive Rate (FPR) were calculated. SHD is defined as the sum of additions, deletions and reversals of edges required to turn one graph into another. In this study, those graphs are the graph output from GES search and the equivalence class of the gold standard graph,  e.g. see \figurename{\ref{Dag2}} for an example of DAG 2. \cite{hauser2012characterization}. Smaller values represent a smaller degree of difference between a found DAG and the equivalence class of the gold standard DAG \cite{tsamardinos2006max}, so lower SHD is preferred. TDR, TPR, FPR, and F1 were computed based upon the presence of an edge alone and did not take edge orientation into account. In contrast, SHD took both edge presence and orientation/directionality into account. 

True Positive Rate (TPR), also commonly referred to as recall or sensitivity, is calculated as the number of correctly found edges in the discovered graph divided by the number of edges in the gold standard DAG from which the data was originally generated. TPR ranges from 0 to 1, with higher values being preferred.

\begin{equation}
    TPR=\frac{True Edges Found in Search}{Total No. Edges in Gold Standard DAG}
\end{equation}

False Positive Rate (FPR) is a Type I error metric. It is calculated by the number of incorrect edges in the discovered graph divided by the number of gaps, or the lack of an edge between any two nodes, in the gold standard graph. FPR ranges from 0 to 1. Lower FPR values are preferred.

\begin{equation}
    FPR=\frac{False Edges}{Gaps in GSDAG}
\end{equation}

True discovery rate (TDR), also commonly referred to as precision or positive predictive value, is defined as the number of correct edges in the discovered graph divided by the number of found edges in the discovered graph. TDR ranges from 0 to 1, with higher values being preferred.
\begin{equation}
    TDR=\frac{TrueEdges}{TotalFoundEdges}
\end{equation}

The F1 score is the harmonic mean between the TPR (recall) and the TDR (precision). It is a performance measure with scores ranging from 0 to 1, with higher values being preferred. An F1 score of 1 corresponds to perfect TPR and TDR. \begin{equation}
F1=2* \frac{TDR*TPR}{TDR+TPR}
\end{equation}

These search performance metrics were calculated for all models, binning conditions, sample sizes, and tuning parameters.

A secondary data simulation was run for each DAG, randomizing  edge parameters of each DAG but maintaining the structure. Edge parameters were again generated ranging from 0.1 to 1, and each variable was given an independent noise term with a Normal distribution. Data was generated by sampling each variable from a distribution equal to the weighted sum of its parents' values in the model, with weights equal to the corresponding edge weights, and its independent noise term. Each edge parameterization condition resulted in only one search data set for each condition, resulting in 15 data sets (5 break conditions by 3 sample size conditions). Searches were run with the same tuning parameters ($\lambda$=1, 2, 4) and performance metrics (SHD, TPR, TDR, FPR, F1).

Plots of results are denoted  as either "Sim1", if from simulation 1 based upon 200 random data sets of the same edge weights, or "Sim 2" if from the secondary analysis based upon 1 randomly generated data set for each of the 200 parameterization of edge weights.

\section{Results}
\label{results}
\begin{figure}[htbp]
\centerline{\includegraphics[width=\linewidth,height=60mm]{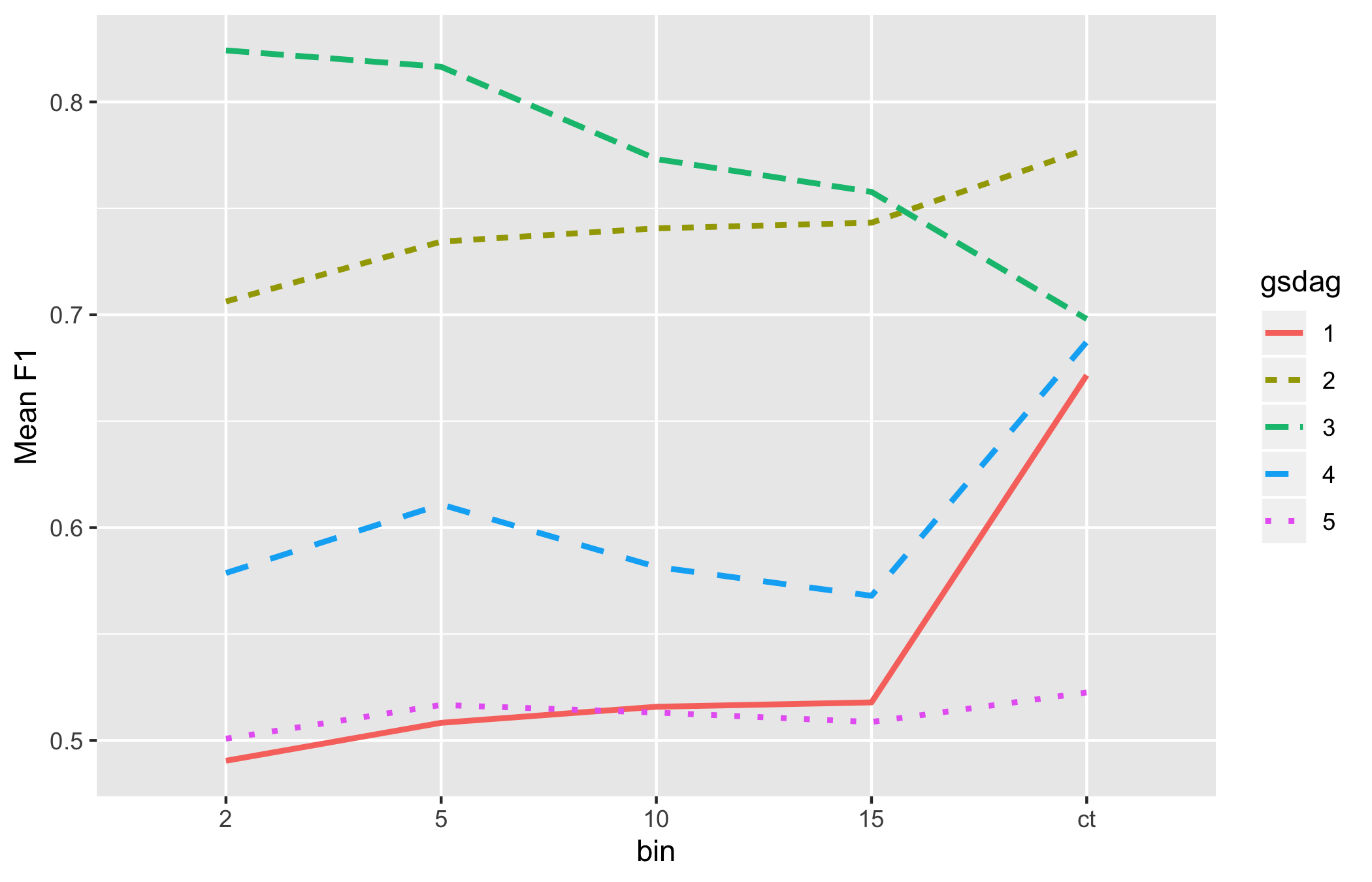}}
\caption{Sim1: F1 by bin condition and DAG}
\label{MainF1Bin}
\end{figure}

Tables located in the online supplement provide summary statistics  for average scores for each condition evaluated and averaged over 200 data sets. Overall performance was measured by mean F1 and SHD. F1 is used as the preferred performance metric when comparing or averaging across graphs due to having a standardized scale, while SHD was used for individual graphs as it accounts for edge orientation. F1 for GES searches of the five DAGs generally fell within the F1=0.5-0.9 range as seen in \figurename{\ref{MainF1Bin}}.

F1 improved with increasing number of bins. GES searches resulted in highest average F1 when provided continuous data, with the exception of a decreased F1 in continuous data for DAG3 searches. GES showed variable performance between the 5 graphs and binning conditions  with lowest average F1's on the densest graph, DAG 5, regardless of binning condition. In contrast, GES searches resulted in higher average F1s in  DAGS 2 and 3. Within DAG 1, F1 was low in binned conditions but improved in the continuous data condition. This can be seen in \figurename{\ref{MainF1Bin}}.

\begin{figure}[htbp]
\centerline{\includegraphics[width=\linewidth,height=60mm]{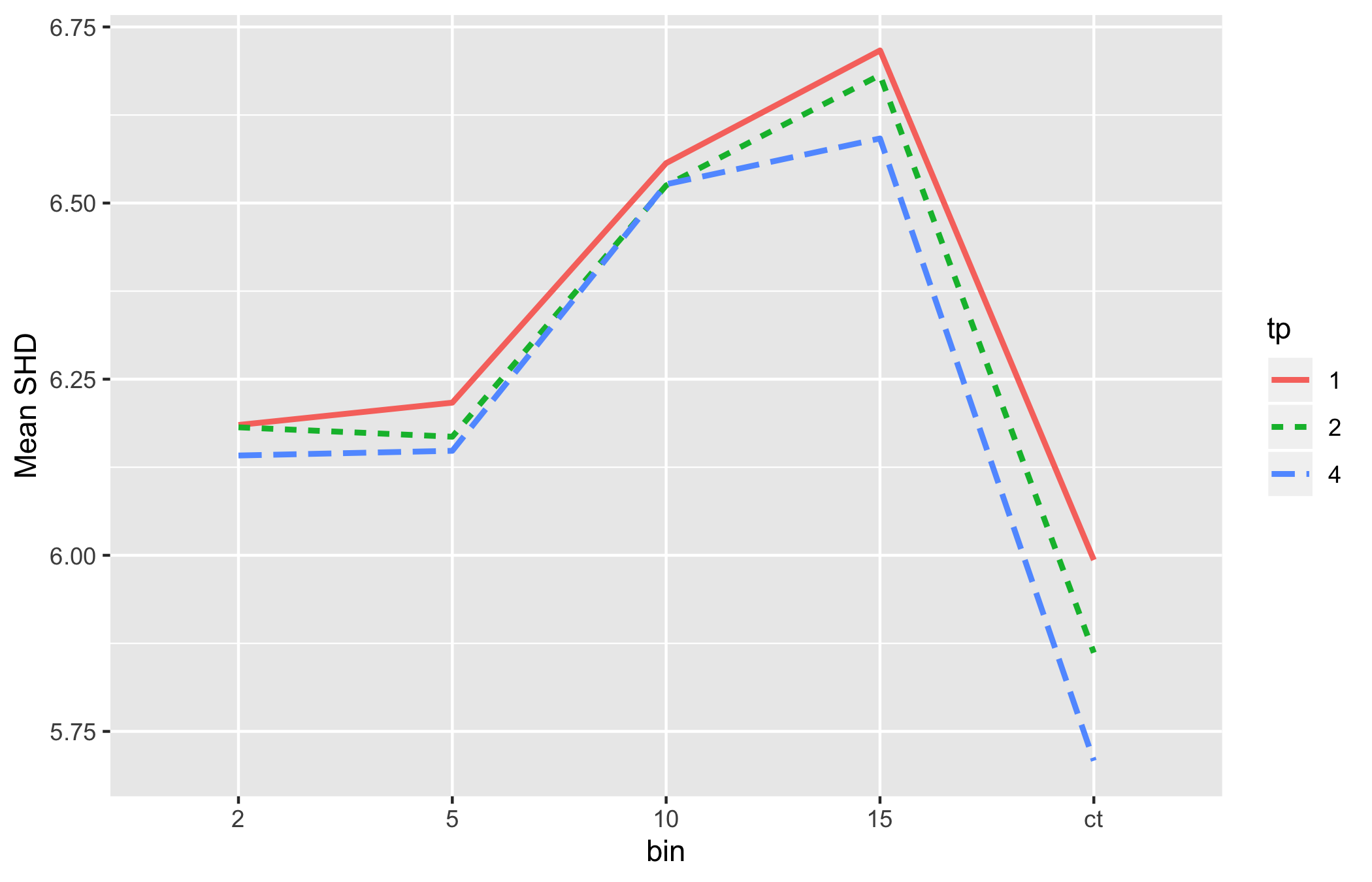}}
\caption{Sim1: SHD by Bin Number and Tuning Parameter in DAG 3}
\label{D3bin}
\end{figure}

\begin{figure}[htbp]
\centerline{\includegraphics[width=\linewidth,height=60mm]{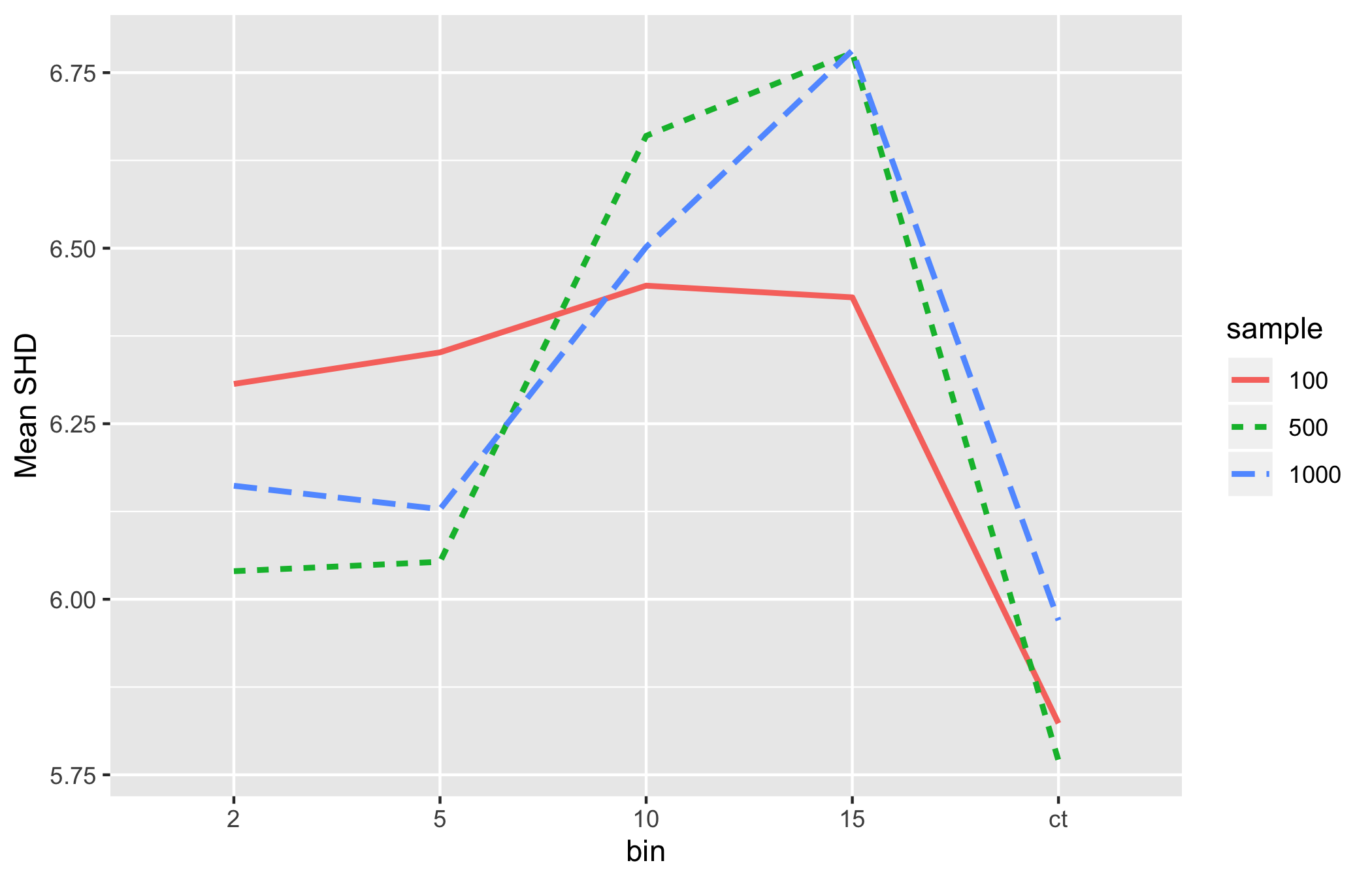}}
\caption{Sim1: SHD by Bin Number and Sample Size in DAG 3}
\label{D3binsam}
\end{figure}

Figures   \ref{D3bin} and \ref{D3binsam} show how SHD changes with sample size and binning condition within DAG 3 searches. In contrast to F1 trends, SHD was lowest in the continuous condition and highest in the 10 and 15 bin conditions, with bin conditions of 2 and 5 showing intermediary SHD score. This trend was seen across all tuning parameters, and to a weaker degree across sample sizes. In the N=100 condition in DAG 3, lower sample size had improved performance at low bin conditions compared to sample sizes of N=500 or N=1000. Within DAG3, tuning parameters choice resulted in relatively similar SHD values.

\begin{figure}[htbp]
\centerline{\includegraphics[width=\linewidth,height=60mm]{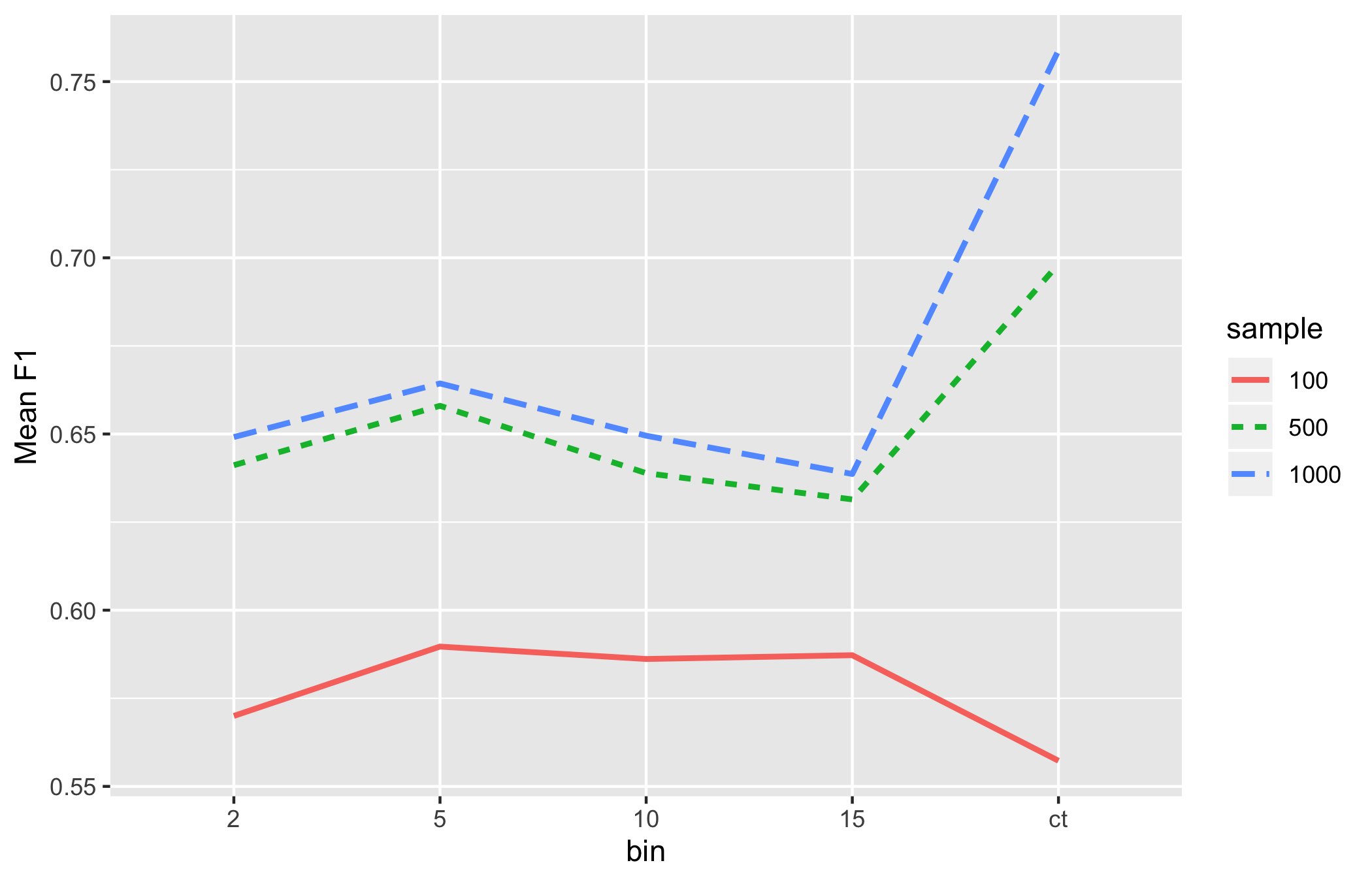}}
\caption{Sim1: Search performance by Bin Number and Observations Across DAGs}
\label{Mainbinsam1}
\end{figure}

Overall (averaged across graphs), F1 increased over sample size and bin condition.  A nonlinear trend of bin condition was observable for all 3 sample sizes. This was most notable within the lowest sample size (N=100) where the highest average F1 occurred in the 5 bin condition and lowest average F1 in the continuous condition. However, at larger sample sizes (N=500, N=1000), continuous data resulted higher F1 scores than binned conditions. \figurename{\ref{Mainbinsam1}}. 

\begin{figure}[htbp]
\centerline{\includegraphics[width=\linewidth,height=60mm]{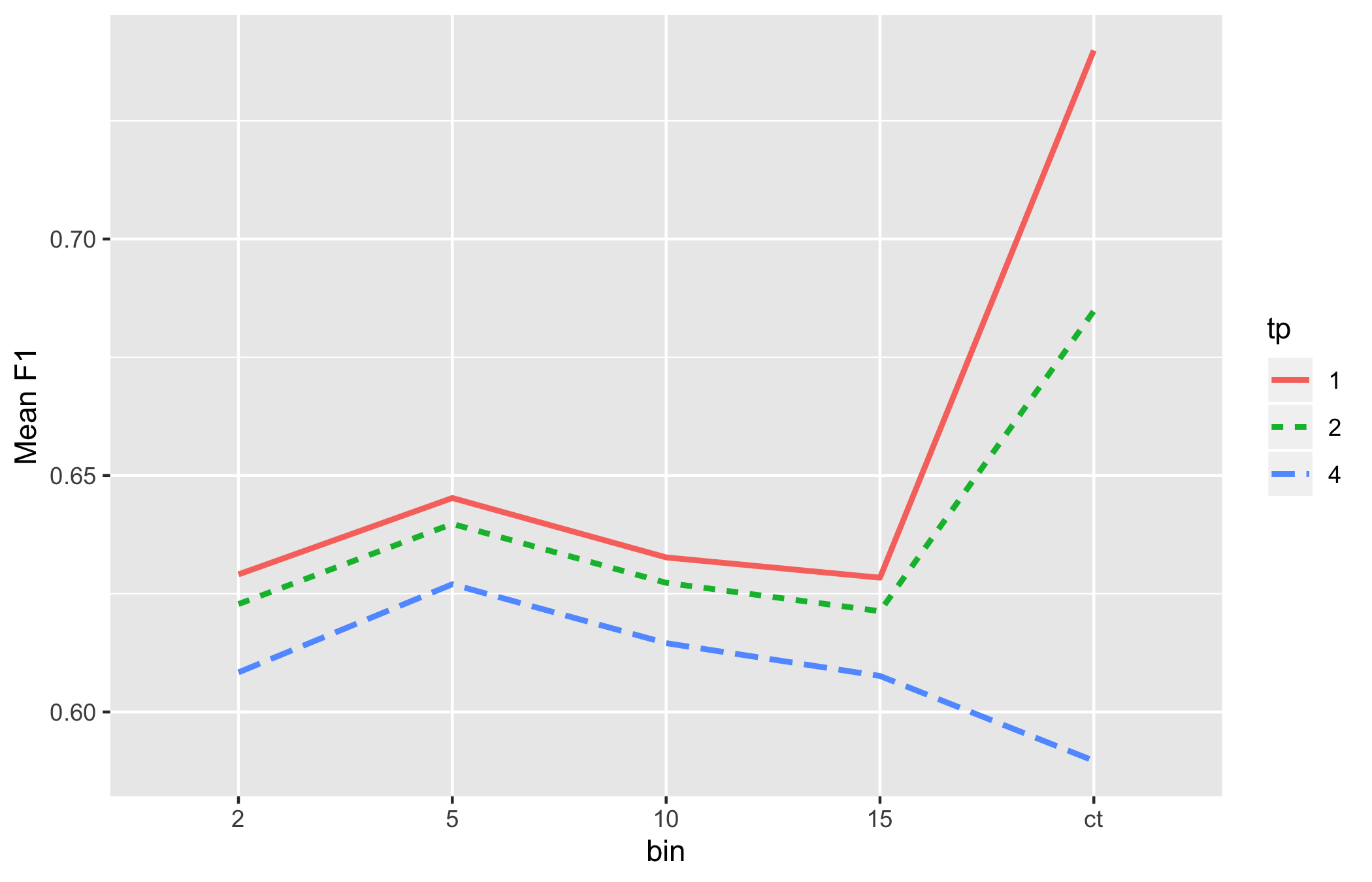}}
\caption{Sim1: F1 by Tuning Parameter and Bin Averaged Across DAGs}
\label{MainF1BinTP}
\end{figure}

\begin{figure}[htbp]
\centerline{\includegraphics[width=\linewidth]{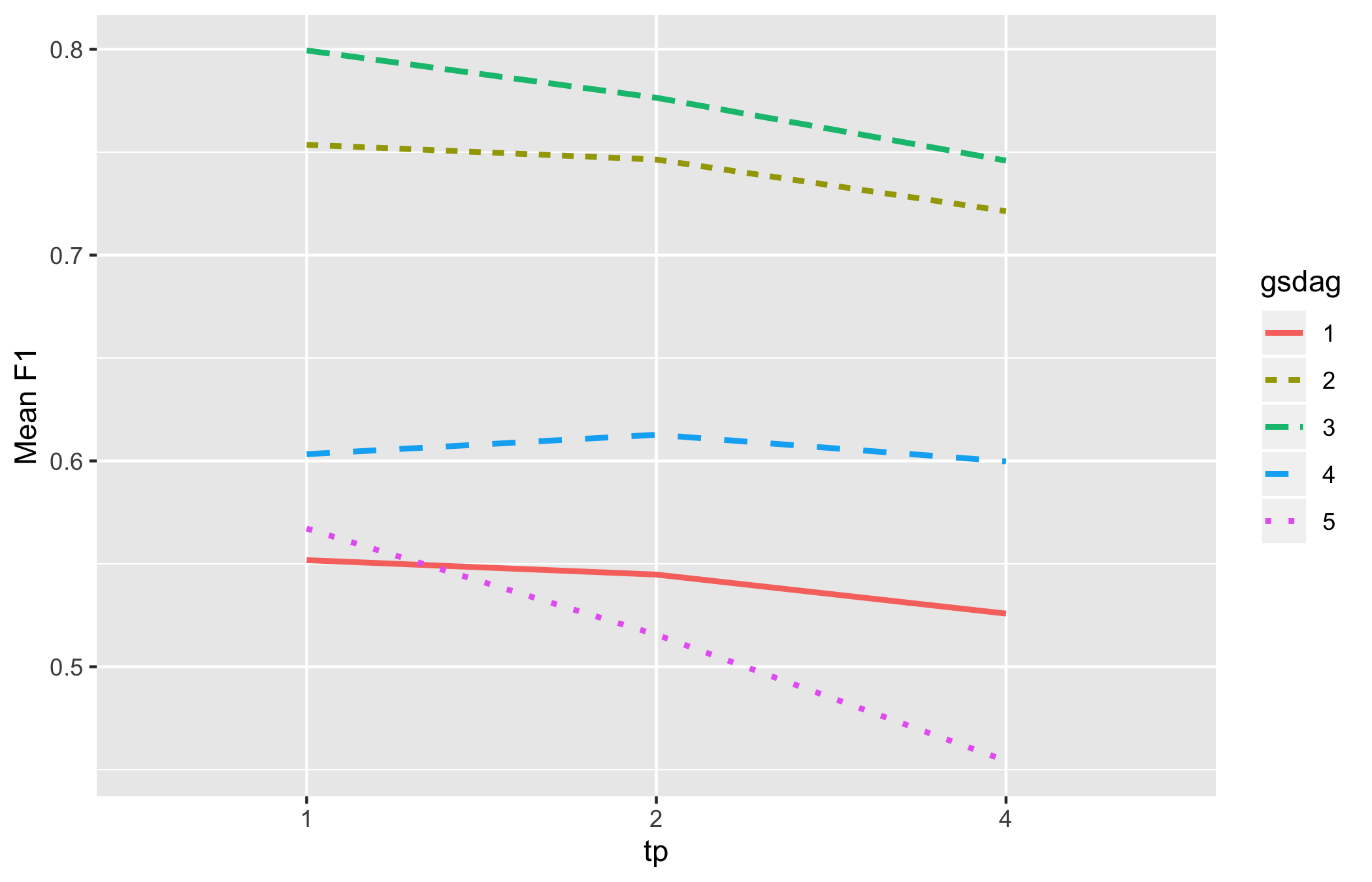}}
\caption{Sim1: Tuning Parameter performance by DAG condition}
\label{MainF1TP}
\end{figure}

As expected, modifying tuning parameters also resulted in differences to  F1 score. Overall, a $\lambda$=1 resulted in highest mean F1 score and $\lambda$=4 lowest average F1 score, as seen in \figurename{\ref{MainF1BinTP}} . However, tuning parameters did not perform uniformly across the 5 graph conditions as seen in \figurename{\ref{MainF1TP}}. The most variable performance was seen in DAG 5, the densest graph, where $\lambda$=1 performed noticeably better than a $\lambda$=4.

\begin{figure}[htbp]
\centerline{\includegraphics[width=\linewidth]{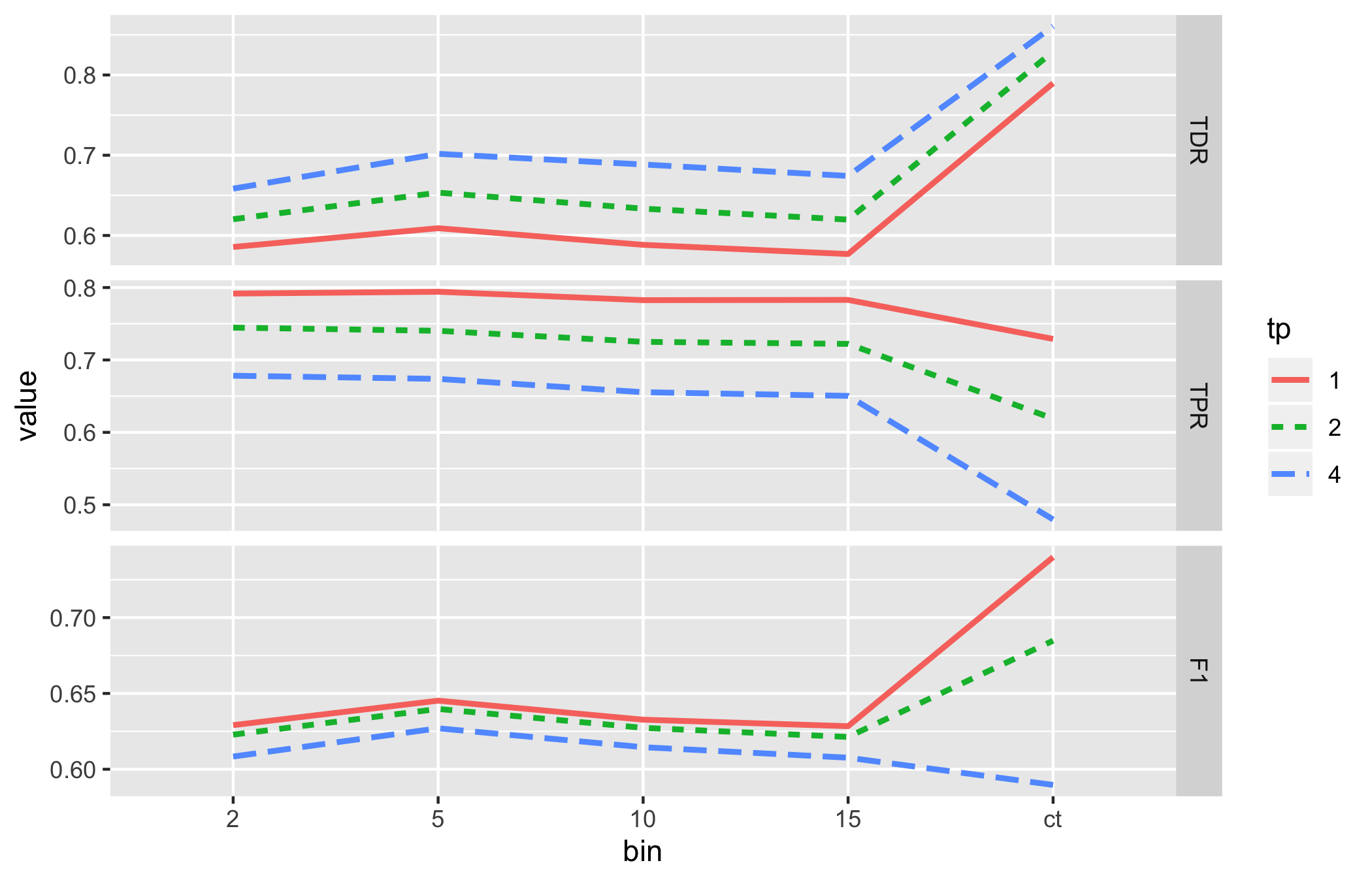}}
\caption{Sim1: Tuning Parameter Performance by Bin Condition}
\label{samptunbin}
\end{figure}

\begin{figure}[htbp]
\centerline{\includegraphics[width=\linewidth]{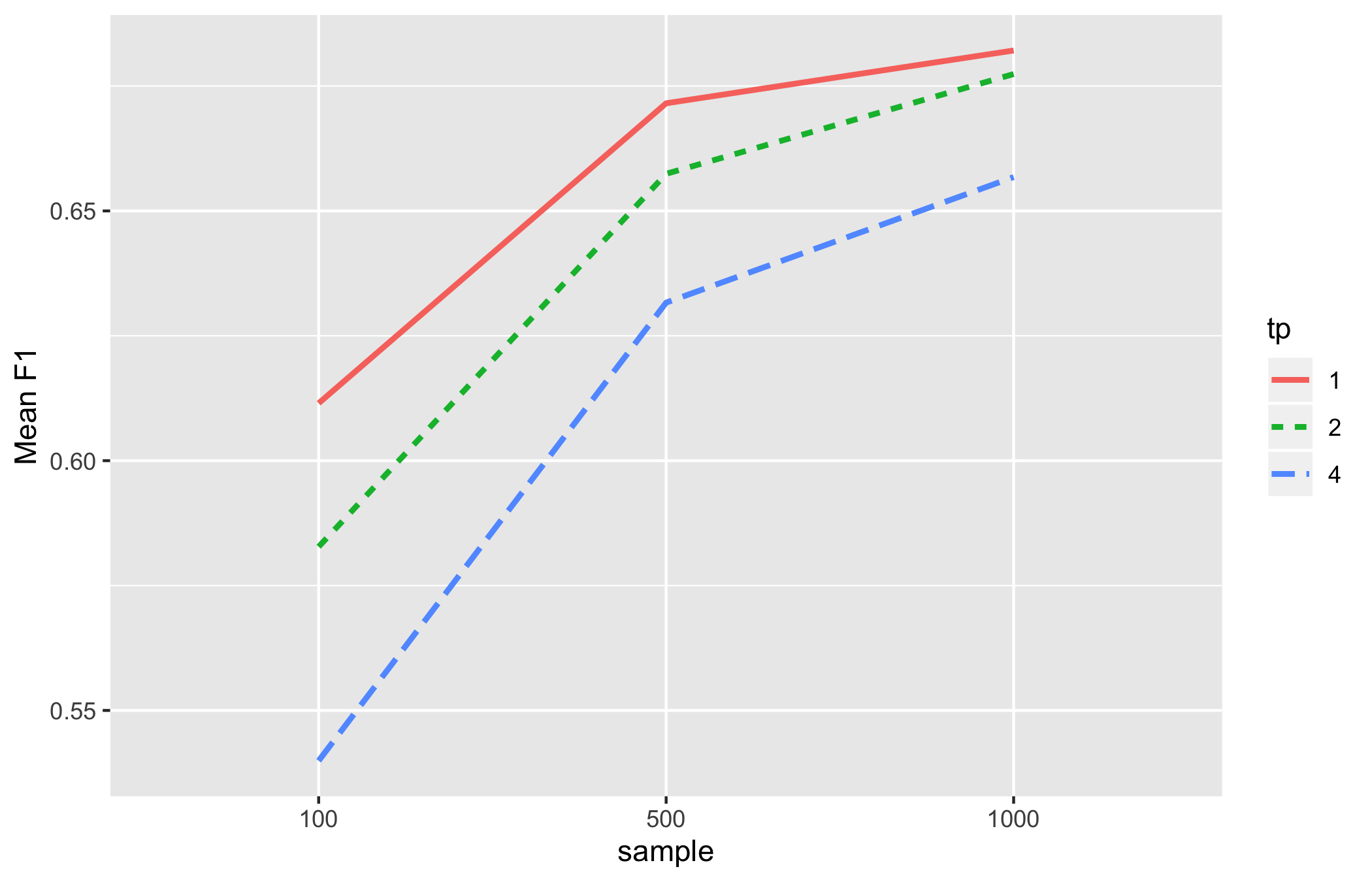}}
\caption{Sim1: Tuning Parameter Performance by Sample}
\label{tpbysam}
\end{figure}

Increasing tuning parameters negatively affected TPR(recall) and positively affected TDR(precision). Tuning parameters had larger effects on lowering TPR than on increasing TDR as seen in
 \figurename{\ref{samptunbin}}. Modifications to tuning parameter $\lambda$ had the most significant effect on F1 when data was continuous \figurename{\ref{Mainbinsam1}} or at lower sample sizes (N=100) \figurename{\ref{tpbysam}}. 

\begin{figure}[htbp]
\centerline{\includegraphics[width=\linewidth,height=60mm]{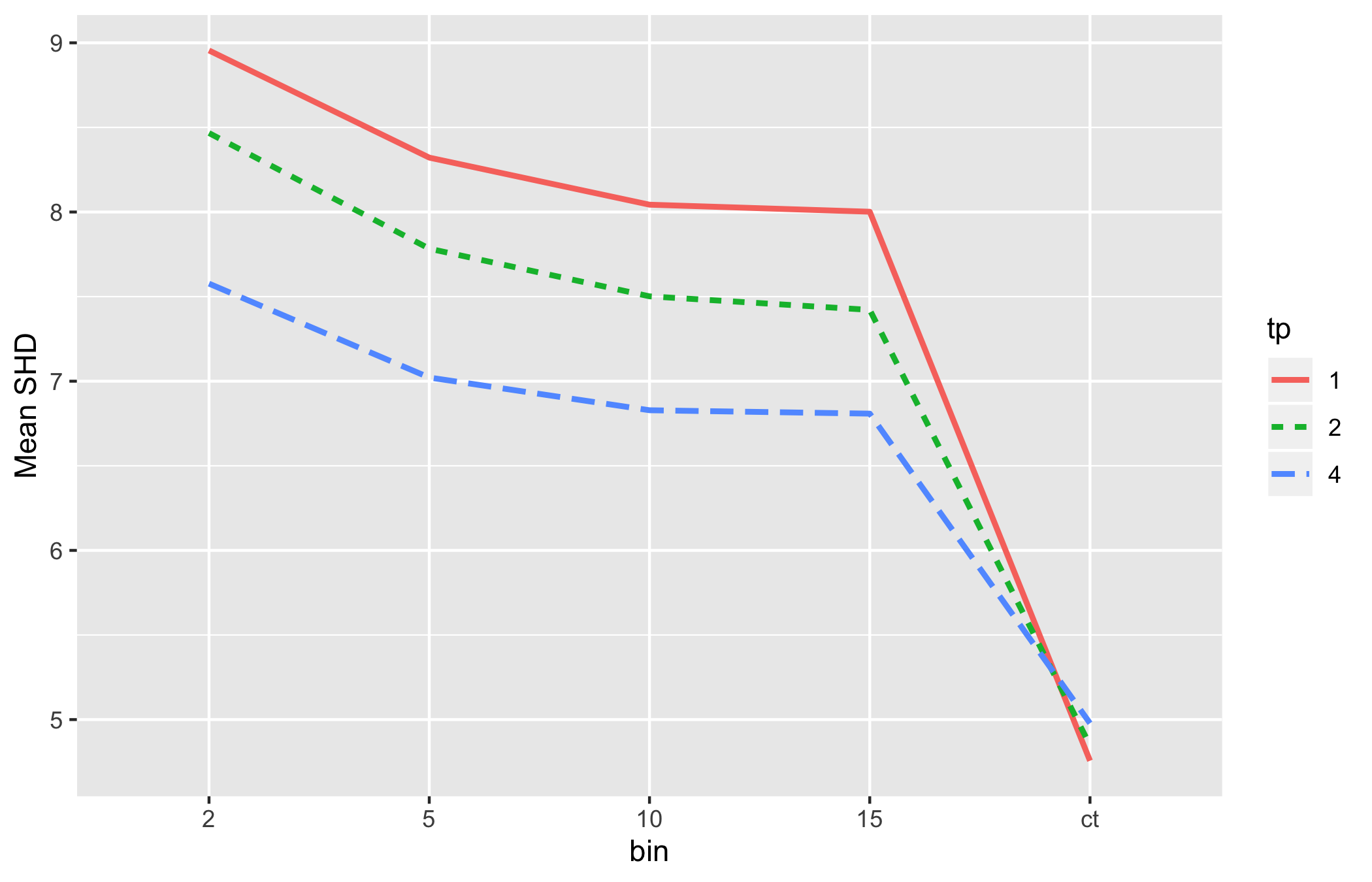}}
\caption{Sim1: SHD by Tuning and Bin Condition in DAG2}
\label{D2F1TPbin}
\end{figure}

\begin{figure}[htbp]
\centerline{\includegraphics[width=\linewidth,height=60mm]{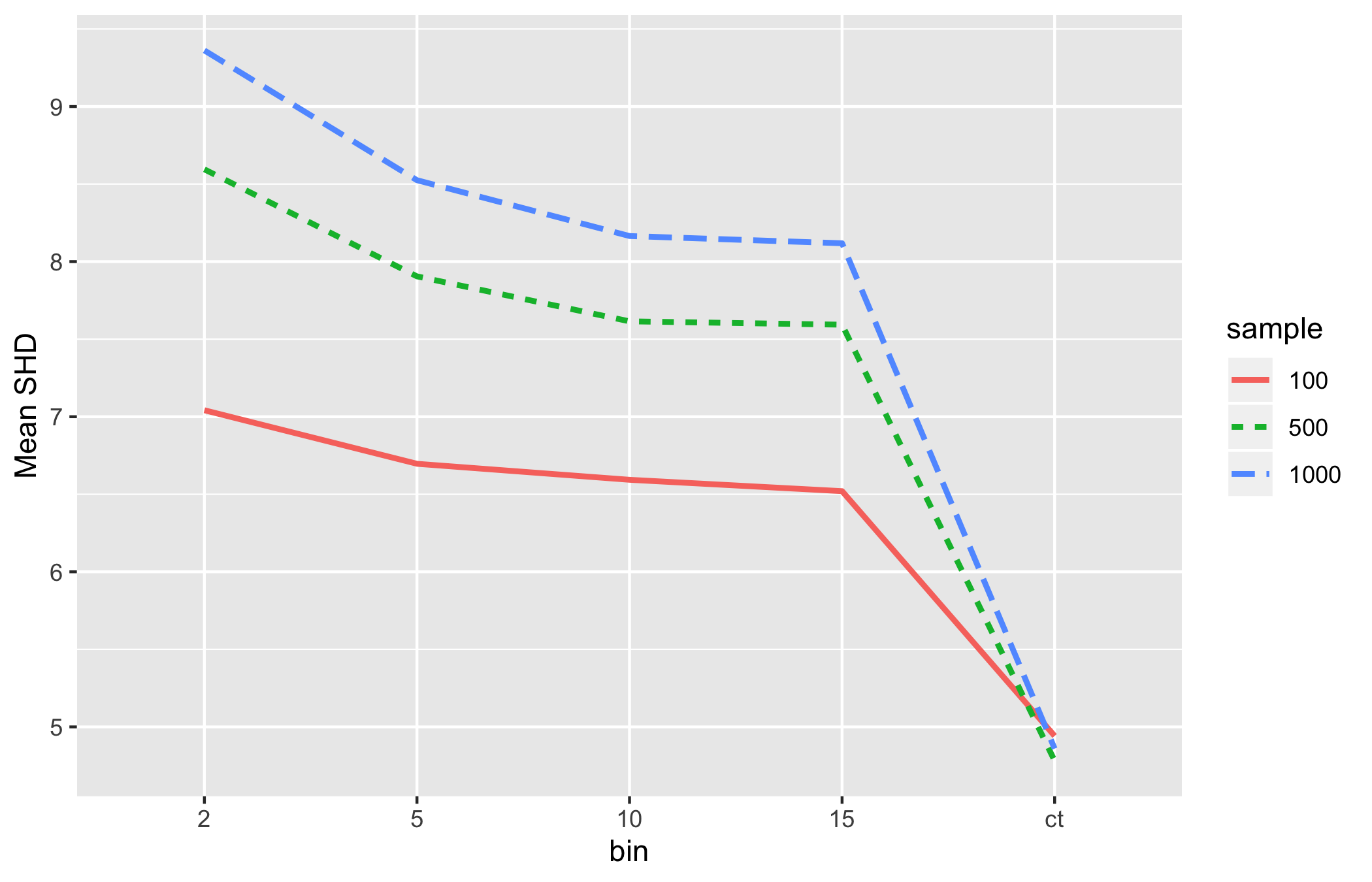}}
\caption{Sim1: SHD by Observations and Bin Condition in DAG2}
\label{D2F1Sbin}
\end{figure}

DAG 2 was selected to run further metrics as it appeared to follow average trends of binning, sample size and tuning parameters on search performance. Within DAG 2, tuning parameter $\lambda=4$  showed lower SHD than $\lambda=2$ and $\lambda=1$, as seen in \figurename{\ref{D2F1TPbin}}. Tuning parameter $\lambda=4$ resulted in the lowest SHD in all binned conditions but not within the the continuous data condition, where it resulted in a negligible increase in SHD. In the continuous condition, tuning parameters performed relatively similar. Similar trends in SHD average were found examining the interaction on sample size and binning condition. Within binned data, changes in sample size resulted in moderate SHD differences, seen in \figurename{\ref{D2F1Sbin}}. In contrast, sample size had little effect on SHD in continuous data. 

As the results from simulation 1, were created from 200 samples of the same edge parameterization, we conducted a secondary analysis to ensure that a specific edge weighting was not driving effects. The results for DAG 2 and 3 are presented below. Results for other DAGS can be found in supplemental material.

\begin{figure}[htbp]
\centerline{\includegraphics[width=\linewidth,height=60mm]{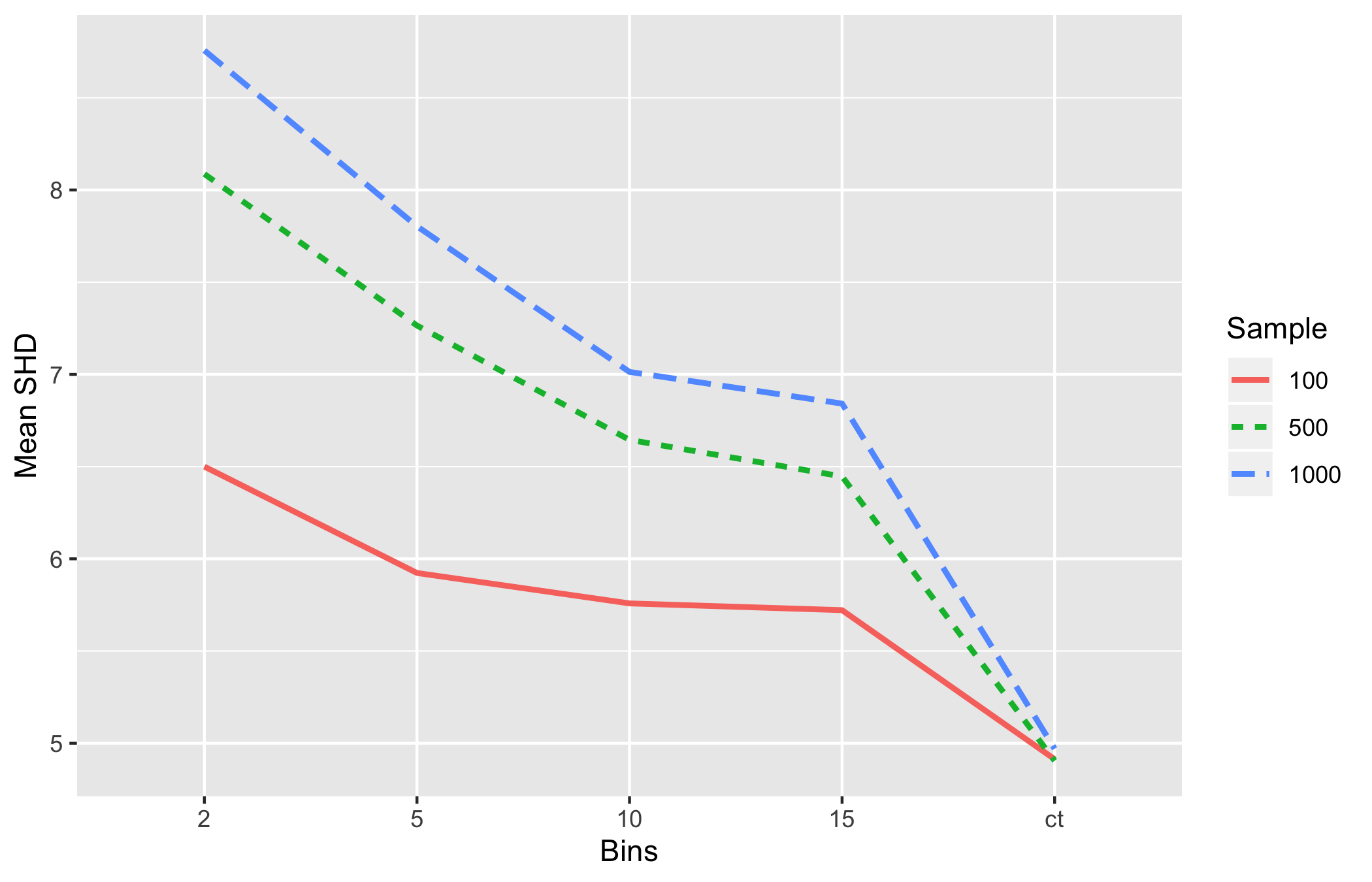}}
\caption{Sim2: Average SHD by Sample and Bin Condition in DAG2}
\label{g2psamBinF1}
\end{figure}

\begin{figure}[htbp]
\centerline{\includegraphics[width=\linewidth,height=60mm]{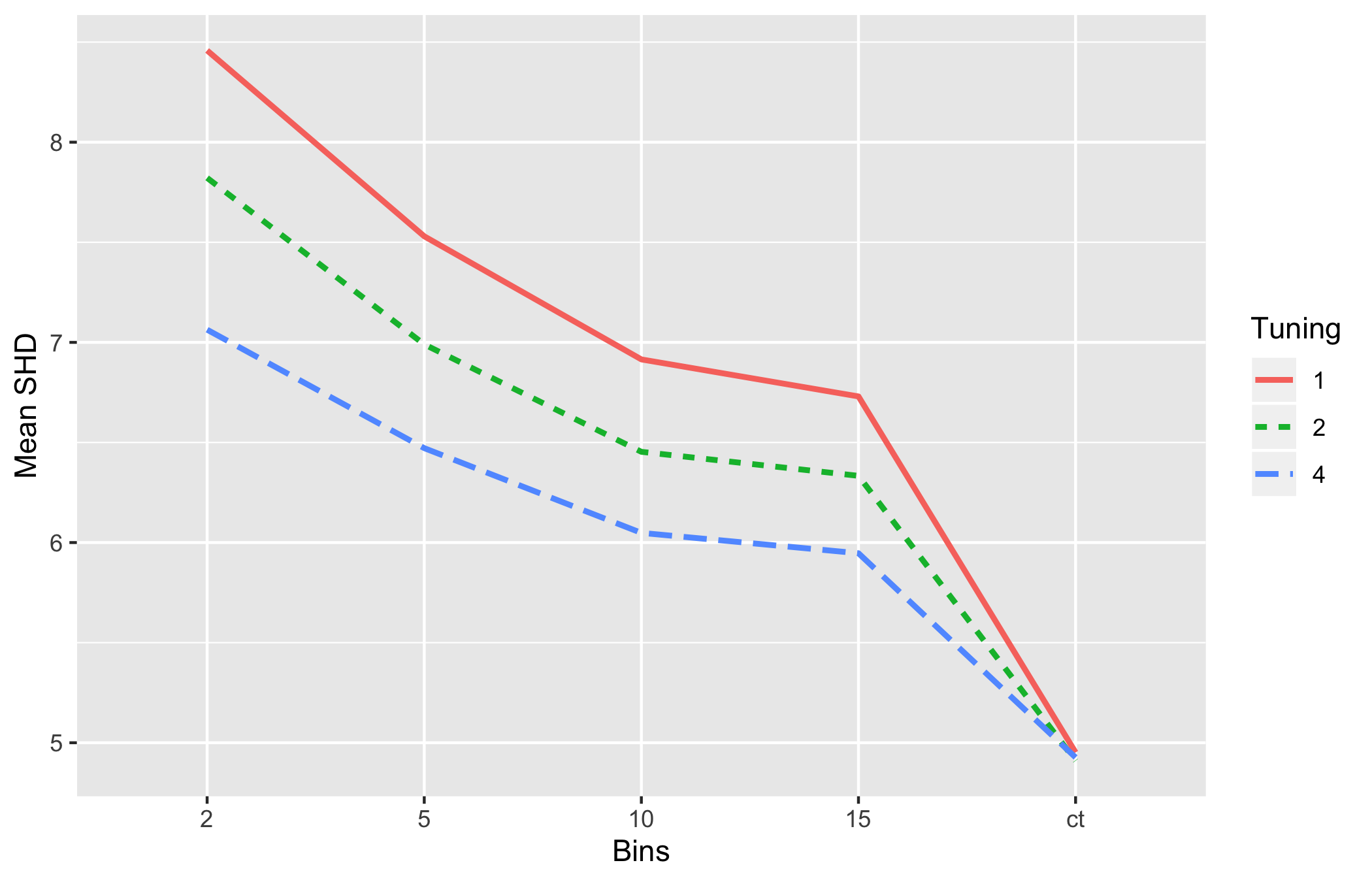}}
\caption{Sim2: SHD by Tuning and Bin Condition in DAG2}
\label{F1D2TPBin}
\end{figure}

The results of the secondary simulation of varying parameterizations of  DAG2 is presented in \figurename{\ref{g2psamBinF1}} and 
\figurename{\ref{F1D2TPBin}}.  Results mirrored findings of the initial simulation, namely lower SHD values with increasing sample size and bin numbers.  Within continuous data, sample size did not appear to affect SHD. However,  SHD values varied by binning condition and by sample size as seen in \figurename{\ref{g2psamBinF1}}. The effect of tuning parameters on SHD, seen in \figurename{\ref{F1D2TPBin}} had similar trends  with $\lambda=4$ resulting in the lowest average SHD and $\lambda=1$ the highest mean SHD. Within the continuous condition, choice of tuning parameters had little effect on SHD. Tuning parameters showed differential effects on SHD within binned data, particularly within low bin conditions. For reference, analogous Simulation 1 results can be seen in \figurename{\ref{D2F1Sbin}} and \figurename{\ref{D2F1TPbin}}. 

\begin{figure}[htbp]
\centerline{\includegraphics[width=\linewidth,height=60mm]{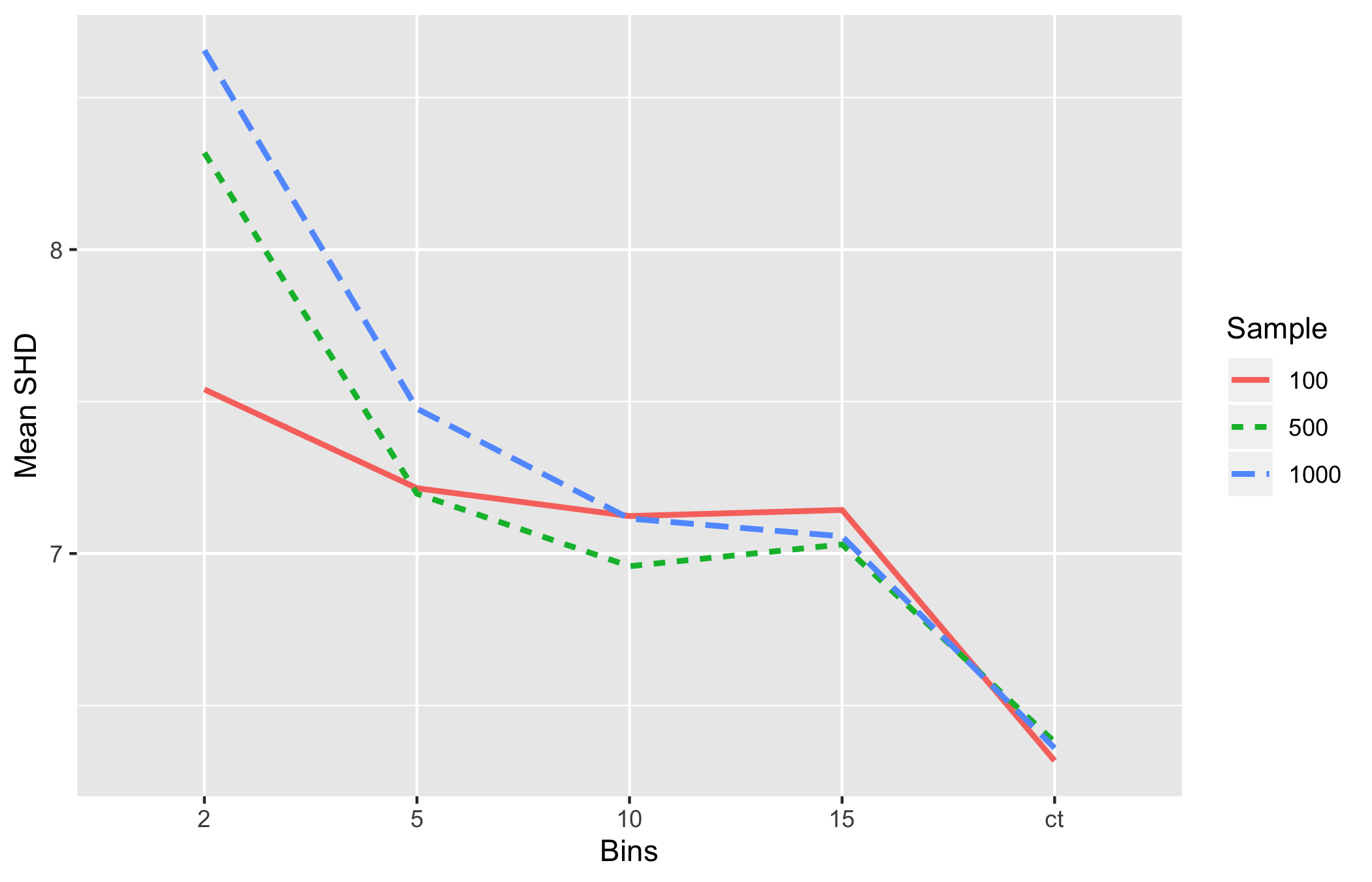}}
\caption{Sim2: Average SHD by Sample and Bin Condition in DAG 3}
\label{G3PSamBinF1}
\end{figure}

\begin{figure}[htbp]
\centerline{\includegraphics[width=\linewidth,height=60mm]{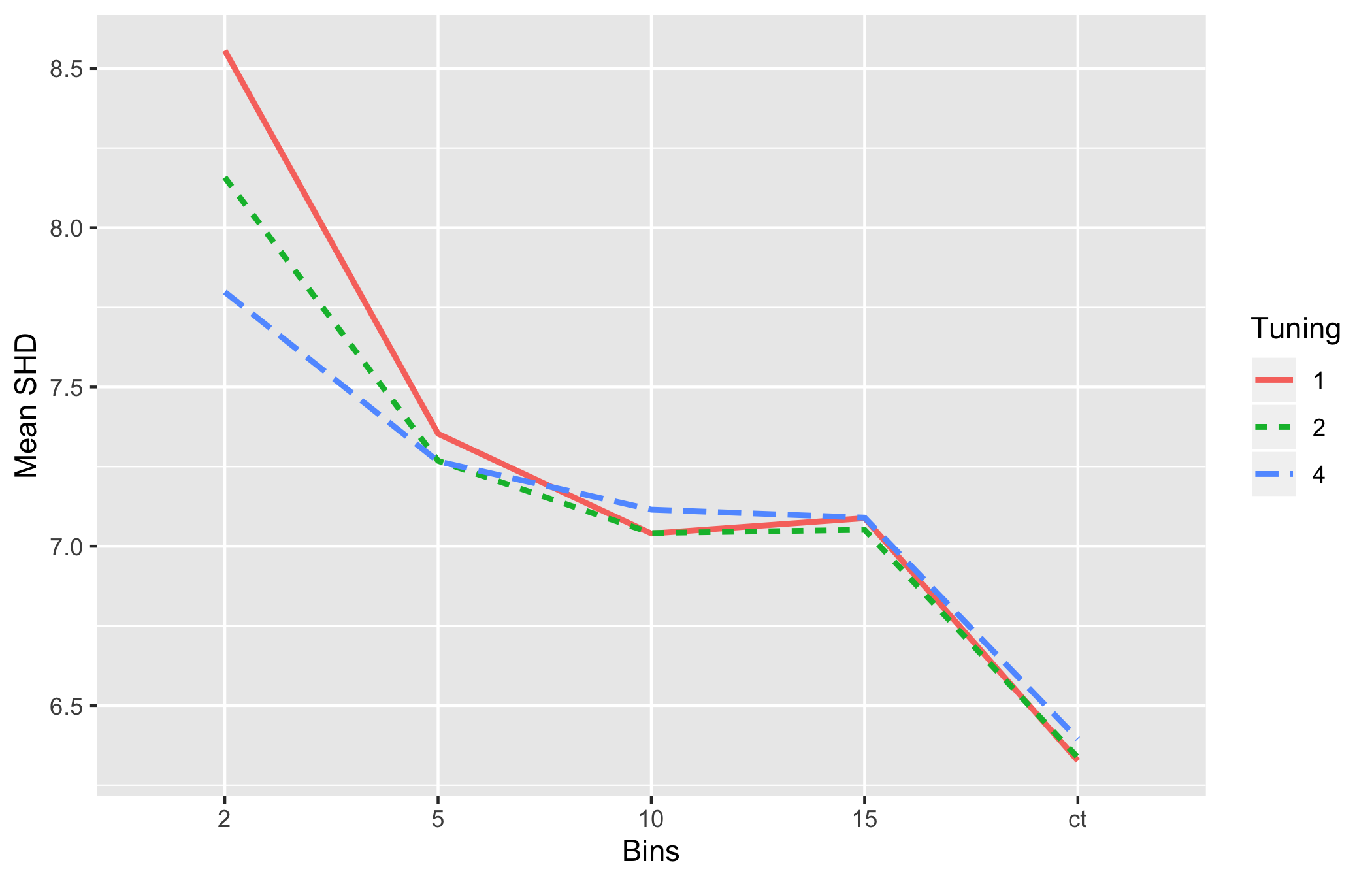}}
\caption{Sim2: Average SHD by Tuning and Bin Condition in DAG3}
\label{MainF1D3TPBin}
\end{figure}

Even with 200 varying edge parameterizations, GES showed variable performance in DAG 3. Continuous data resulted in the lowest average SHD and the 2 bin condition, the highest SHD, see \figurename{\ref{G3PSamBinF1}}. However, effects of sample size and tuning parameter on SHD were less clear cut, particularly in the 5 bin, 10 bin, and 15 bin conditions. For example, the lowest average SHD in the 2 and 5 bin conditions occurred in the sample sizes of N=100. While tuning parameter choice appeared to effect SHD in the 2 bin condition, choice of tuning parameter resulted in minimal SHD differences in conditions of 5 or more bins, see \figurename{\ref{MainF1D3TPBin}}. For comparison, original simulation results can be seen in \figurename{\ref{D3binsam}} and \figurename{\ref{D3bin}}.

\begin{figure}[htbp]
\centerline{\includegraphics[width=\linewidth,height=60mm]{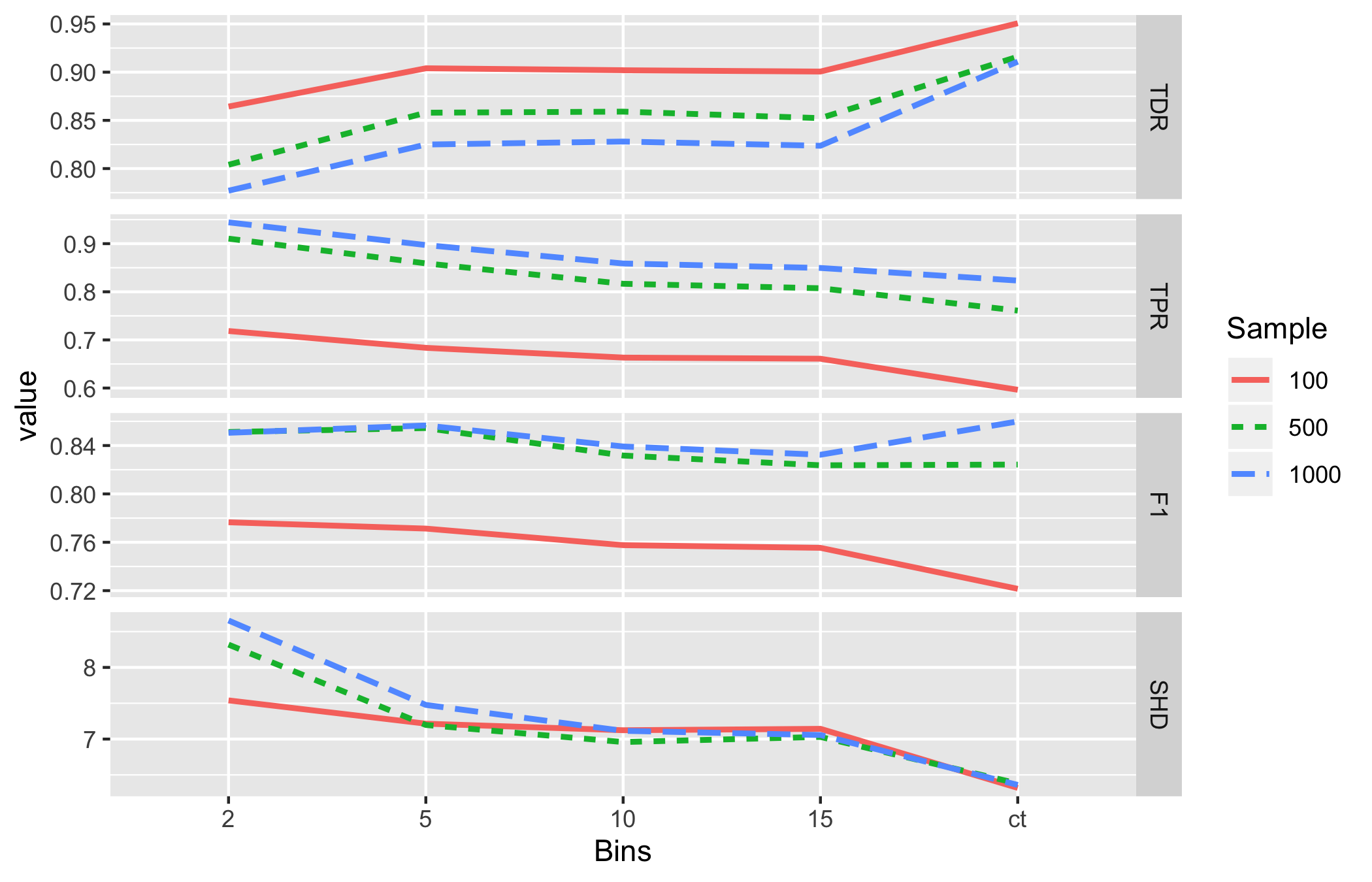}}
\caption{Sim2: Performance Metrics Across Tuning and Bin Condition in DAG 3}
\label{3samparmetDG3}
\end{figure}

\section{Discussion}
\label{discussion}

Perhaps the most striking finding is the lack of any hard and fast rules for universally improving search performance, although general trends and rules emerged in both simulations. GES had lowest SHD, when given continuous data.  Of note, search performance of GES differed depending on whether edge orientations were taken into account, with SHD accounting for directionality in contrast to other performance metrics which reflected only adjacencies. F1 showed an unusual trend of increasing as as bins increased within DAG 3. In contrast, SHD had a nonlinear effect, with lowest SHD in the continuous data and intermediary valued SHD in lower binned data. DAG 3 appears to represent at least one case in which using continuous data would result in poor relative performance of adjacency discovery alone. This was seen to a lesser degree in the secondary analysis suggesting that edge weights may influence the degree of this effect.The secondary simulation study also showed a general trend of SHD decreasing with increasing bin condition.

As only 1 DAG was generated in each gold standard condition, it is difficult to characterize the number and type of graphs in which this paradoxical decrease in F1 with increasing is present. It is possible that this is a result of the specific structure of DAG 3, and is unlikely to be seen in other DAGs of similar edge probability and/or node number. Of note, DAG 3 was characterized by a 4-clique with a separate unrelated parent to the sink node, see \figurename{\ref{Dag3}}. The use of only one structure for each gold standard graph is a significant limitation of these findings. Future work should examine performance in a broader range of randomly generated DAGs of similar node size and density.

Increasing sample size led to increased F1 with N=100 having the lowest F1 values. In samples of N=500 or N=1000, F1 was relatively comparable, except in continuous data where sample size N=1000 showed a higher F1.   Tuning parameters showed similar trends across binning conditions. With the default tuning parameter $\lambda=1$  resulting in the highest performance when examining adjacency structure only (F1) but the lowest when accounting for edge direction (SHD). Modifying $\lambda$ resulted in small changes to F1 in the binned condition, but noticeable differences in the continuous condition. This suggests that tuning parameters need to be carefully chosen for continuous data  but may be less important for binned data in terms of F1. SHD, which accounted for edge direction, showed a divergent pattern within DAG 2 and DAG 3 showing the highest variance in the binned conditions. However, it is premature to label this incongruence as a rule without examining further graphs and parameterizations. Tuning parameters performed differentially across the five DAGs, although altering tuning parameter resulted in only small changes to F1 averages within any given DAG. Tuning parameters had a larger effect on the TPR/recall  than on TDR/precision. Some DAGs appeared more sensitive to poorly chosen tuning parameters than others (e.g. DAG 5). Overall, results suggest GES produces the highest but also the most variable F1 by sample and tuning parameter in continuous data. Within DAG 2 and DAG 3, SHD showed higher sensitive  to sample size and tuning parameter in binned conditions than continuous data. Tuning parameter and sample size had little effect on SHD in the continuous condition. Among binned conditions, 5 to 10 bins appeared to optimize F1 slightly. 

A significant limitation of this research, as mentioned previously, was the simulation of only one DAG in each graph size and density condition. Future research should generate several structures of similar node size, edge number, and density to validate these findings across differing structures and parameterizations. Future research may benefit from examining differences in search algorithm performance on binned data as these findings may not generalize to other algorithms. They also may not generalize to data with missing values, with non-Gaussian independent noise terms, or with unmeasured common causes. Additional studies should be performed to evaluate the effect of binning, tuning parameters, and other user-controllable search conditions in these other settings.

Despite these limitations, the results have important implications for those using GES in their research. Researchers should be aware that no hard and fast rule to maximize performance based upon sample size, number of bins or tuning parameter exists. Choice of tuning parameters and sample sizes could result in notable differences in performance, but did not necessarily lead to differences in search performance. While continuous data results in the best search performance overall, it could also result in the worst performance, depending on the data generating model and chosen search performance metric.

\bibliography{IEEEabrv,5194Paper}{}
\bibliographystyle{IEEEtran}

\end{document}